\documentclass[runningheads]{llncs}
\usepackage{graphicx}
\usepackage{cite}
\usepackage{amsmath,amssymb,amsfonts}
\usepackage{algorithmic}
\usepackage{graphicx}
\usepackage{textcomp}
\usepackage{adjustbox}
\usepackage{multirow}
\usepackage{bbold}
\usepackage{siunitx}
\usepackage{hyperref}       
\usepackage{xcolor}
\usepackage{sidecap}

\usepackage{booktabs}
\usepackage[mathscr]{eucal}

\hypersetup{
    colorlinks,
    linkcolor={blue!50!black},
    citecolor={blue!50!black},
    urlcolor={blue!50!black}
}
\usepackage{amsmath, bm}
\usepackage{hyperref}

\usepackage{pifont}

\usepackage{floatrow}
\newfloatcommand{capbtabbox}{table}[][\FBwidth]

\begin{document}

\title{Rectifying Noisy Labels with Sequential Prior: Multi-Scale Temporal Feature Affinity Learning for Robust Video Segmentation}
\author{Beilei Cui\inst{*,1}\orcidID{0009-0009-7900-8032} \and
Minqing Zhang\inst{*,2}\orcidID{0000-0002-7214-0569} \and
Mengya Xu\inst{*,3}\orcidID{0000-0002-4338-7079} \and
An Wang\inst{1}\orcidID{0000-0001-5515-0653} \and
Wu Yuan\inst{\dag,2}\orcidID{0000-0001-9405-519X} \and
Hongliang Ren\inst{\dag,1,3}\orcidID{0000-0002-6488-1551}}

\authorrunning{Cui et al.}
%
\institute{Dept. of Electronic Engineering, The Chinese University of Hong Kong, Hong Kong SAR, China \and
Dept. of Biomedical Engineering, The Chinese University of Hong Kong, Hong Kong SAR, China \and 
Dept. of Biomedical Engineering, National University of Singapore, Singapore \\
\email{beileicui@link.cuhk.edu.hk, lars.zhang@link.cuhk.edu.hk, mengya@u.nus.edu, wa09@link.cuhk.edu.hk, wyuan@cuhk.edu.hk, ren@nus.edu.sg }}

\maketitle              
\begin{abstract}

Noisy label problems are inevitably in existence within medical image segmentation causing severe performance degradation. Previous segmentation methods for noisy label problems only utilize a single image while the potential of leveraging the correlation between images has been overlooked. Especially for video segmentation, adjacent frames contain rich contextual information beneficial in cognizing noisy labels. Based on two insights, we propose a Multi-Scale Temporal Feature Affinity Learning (MS-TFAL) framework to resolve noisy-labeled medical video segmentation issues. First, we argue the sequential prior of videos is an effective reference, i.e., pixel-level features from adjacent frames are close in distance for the same class and far in distance otherwise. Therefore, Temporal Feature Affinity Learning (TFAL) is devised to indicate possible noisy labels by evaluating the affinity between pixels in two adjacent frames. We also notice that the noise distribution exhibits considerable variations across video, image, and pixel levels. In this way, we introduce Multi-Scale Supervision (MSS) to supervise the network from three different perspectives by re-weighting and refining the samples. This design enables the network to concentrate on clean samples in a coarse-to-fine manner. Experiments with both synthetic and real-world label noise demonstrate that our method outperforms recent state-of-the-art robust segmentation approaches. Code is available at \url{https://github.com/BeileiCui/MS-TFAL}.

\end{abstract}

\renewcommand{\thefootnote}{}
\footnotetext{\inst{*} Authors contributed equally to this work.}
\footnotetext{\inst{\dag} Corresponding Author.}

\section{Introduction}

Video segmentation, which refers to assigning pixel-wise annotation to each frame in a video, is one of the most vital tasks in medical image analysis. Thanks to the advance in deep learning algorithms based on Convolutional Neural Networks, medical video segmentation has achieved great progress over recent years~\cite{LITJENS201760}. But a major problem of deep learning methods is that they are largely dependent on both the quantity and quality of training data~\cite{dlsurvey}. Datasets annotated by non-expert humans or automated systems with little supervision typically suffer from very high label noise and are extremely time-consuming. Even expert annotators could generate different labels based on their cognitive bias~\cite{karimi2020deep}. Based on the above, noisy labels are inevitably in existence within medical video datasets causing misguidance to the network and resulting in severe performance degradation. Hence, it is of great importance to design medical video segmentation methods that are robust to noisy labels within training data~\cite{guo2022joint,zhang2020characterizing}.

Most of the previous noisy label methods mainly focus on classification tasks. Only in recent years, the problem of noise labels in segmentation tasks has been more explored, but still less involved in medical image analysis. Previous techniques for solving noisy label problems in medical segmentation tasks can be categorized in three directions. The first type of method aims at deriving and modeling the general distribution of noisy labels in the form of Noise Transition Matrix (NTM)~\cite{pmlr-v139-li21l, guo2021metacorrection}. Secondly, some researchers develop special training strategies to re-weight or re-sample the data such that the model could focus on more dependable samples. Zhang et al.~\cite{zhang2020robust} concurrently train three networks and each network is trained with pixels filtered by the other two networks. Shi et al.~\cite{shi2021distilling} use stable characteristics of clean labels to estimate samples' uncertainty map which is used to further guide the network. Thirdly, label refinement is implemented to renovate noisy labels. Li et al.~\cite{li2021superpixel} represent the image with superpixels to exploit more advanced information in an image and refine the labels accordingly. Liu et al.~\cite{liu2021s} use two different networks to jointly determine the error sample, and use each other to refine the labels to prevent error accumulation. Xu et al.~\cite{xu2022anti} utilize the mean-teacher model and Confident learning to refine the low-quality annotated samples.

Despite the amazing performance in tackling noisy label issues for medical image segmentation, almost all existing techniques only make use of the information within a single image. \emph{To this end, we make the effort in exploring the feature affinity relation between pixels from consecutive frames.} The motivation is that the embedding features of pixels from adjacent frames should be close if they belong to the same class, and should be far if they belong to different classes. Hence, if a pixel's feature is far from the pixels of the same class in the adjacent frame and close to the ones of different classes, its label is more likely to be incorrect. Meanwhile, the distribution of noisy labels may vary among different videos and frames, which also motivates us to supervise the network from multiple perspectives.

Inspired by the motivation above and to better resolve noisy label problems with temporal consistency, we propose Multi-Scale Temporal Feature Affinity Learning (MS-TFAL) framework. Our contributions can be summarized as the following points:

\begin{enumerate}
    \item In this work, we first propose a novel Temporal Feature Affinity Learning (TFAL) method to evaluate the temporal feature affinity map of an image by calculating the similarity between the same and different classes' features of adjacent frames, therefore indicating possible noisy labels.
    
    \item We further develop a Multi-Scale Supervision (MSS) framework based on TFAL by supervising the network through video, image, and pixel levels. Such a coarse-to-fine learning process enables the network to focus more on correct samples at each stage and rectify the noisy labels, thus improving the generalization ability.
    
    \item  Our method is validated on a publicly available dataset with synthetic noisy labels and a real-world label noise dataset and obtained superior performance over other state-of-the-art noisy label techniques.
    
    \item   To the best of our knowledge, we are the first to tackle noisy label problems using inter-frame information and discover the superior ability of sequential prior information to resolve noisy label issues. 
\end{enumerate}

\section{Method}

\begin{figure}[t]
\centering
\includegraphics[width=1\linewidth]{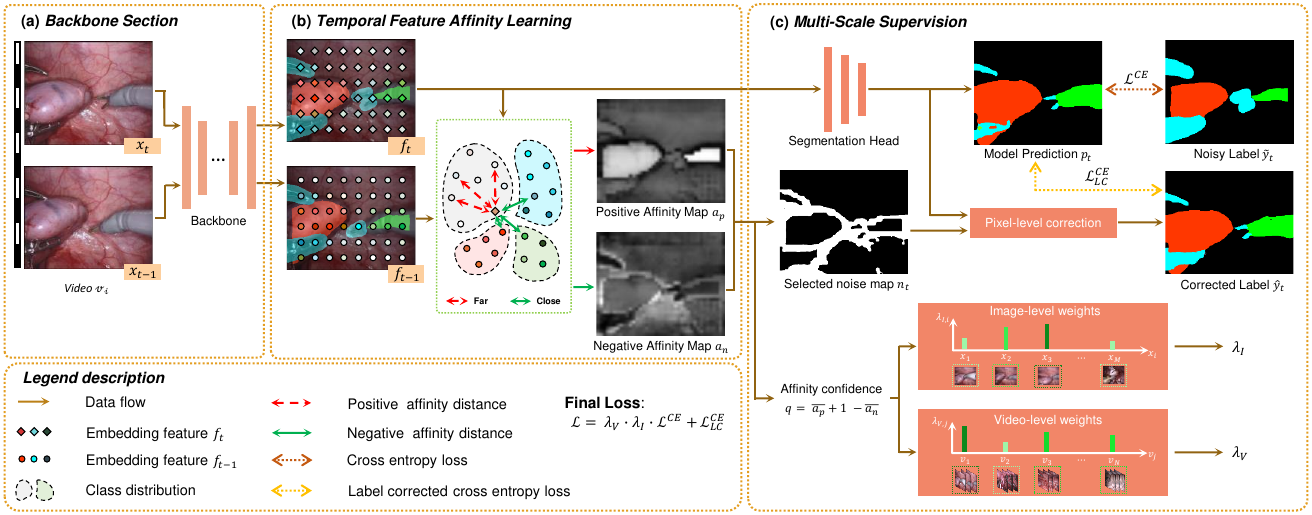}
\caption{Illustration of proposed Multi-Scale Temporal Feature Affinity Learning framework. We acquire the embedding feature maps of adjacent frames in the Backbone Section. Then, the temporal affinity is calculated for each pixel in current frame to obtain the positive and negative affinity map indicating possible noisy labels. The affinity maps are then utilized to supervise the network in a multi-scale manner.}
\label{fig:model}
\end{figure}

The proposed Multi-Scale Temporal Feature Affinity Learning Framework is illustrated in Fig.~\ref{fig:model}. We aim to exploit the information from adjacent frames to identify the possible noisy labels, thereby learning a segmentation network robust to label noises by re-weighting and refining the samples. Formally, given an input training image $x_{t}\in \mathbb{R}^{H\times W\times 3}$, and its adjacent frame $x_{t-1}$, two feature maps $f_{t}, f_{t-1}\in \mathbb{R}^{h\times w\times C_{f}}$ are first generated by a CNN backbone, where $h$,$w$ and $C_{f}$ represent the height, width and channel number. Intuitively, for each pair of features from $f_{t}$ and $f_{t-1}$, their distance should be close if they belong to the same class and far otherwise. Therefore for each pixel in $f_{t}$, we calculate two affinity relations with $f_{t-1}$. The first one is called positive affinity, computed by averaging the cosine similarity between one pixel $f_{t}\left (  i\right )$ in the current frame and all the same class' pixels as $f_{t}\left (  i\right )$ in previous frame. The second one is called negative affinity, computed by averaging the cosine similarity between one pixel $f_{t}\left (  i\right )$ in current frame and all the different class' pixels as $f_{t}\left (  i\right )$ in previous frame. Then through up-sampling, the Positive Affinity Map $a_{p}$ and Negative Affinity Map $a_{n}$ can be obtained, where $a_{p},a_{n}\in  \mathbb{R}^{H\times W}$, denote the affinity relation between $x_{t}$ and $x_{t-1}$. The positive affinity of clean labels should be high while the negative affinity of clean labels should be low. Therefore, the black areas in $a_{p}$ and the white areas in $a_{n}$ are more likely to be noisy labels.

Then we use two affinity maps $a_{p},a_{n}$ to conduct Multi-Scale Supervision training. Multi-scale refers to video, image, and pixel levels. Specifically, for pixel-level supervision, we first obtain thresholds $t_{p}$ and $t_{n}$ by calculating the average positive and negative affinity over the entire dataset. The thresholds are used to determine the possible noisy label sets based on positive and negative affinity separately. The intersection of two sets is selected as the final noisy set and relabeled with the model prediction $p_{t}$. The affinity maps are also used to estimate the image-level weights $\lambda _{I}$ and video-level weights $\lambda _{V}$. The weights enable the network to concentrate on videos and images with higher affinity confidence. Our method is a plug-in module that is not dependent on backbone type and can be applied to both image-based backbones and video-based backbones by modifying the shape of inputs and feature maps. 

\subsection{Temporal Feature Affinity Learning}

The purpose of this section is to estimate the affinity between pixels in the current frame and previous frame, thus indicating possible noisy labels. Specifically, in addition to the aforementioned feature map $f_{t}, f_{t-1}\in \mathbb{R}^{h\times w\times C_{f}}$, we obtain the down-sampled labels with the same size of feature map $\tilde{y}_{t}^{'},\tilde{y}_{t-1}^{'}\in \mathbb{R}^{h\times w\times \mathscr{C}}$, where $\mathscr{C}$ means the total class number. We derive the positive and negative label maps with binary variables: $M_{p},M_{n}\subseteq  \left\{ 0,1\right\}^{hw\times hw}$. The value corresponds to pixel $\left (  i,j\right )$ is determined by the label as:

\begin{equation}
M_{p}\left (  i,j\right ) = \mathbb{1}\left [ \tilde{y}_{t}^{'}\left ( i \right ) = \tilde{y}_{t-1}^{'}\left ( j \right ) \right ],\quad \quad M_{n}\left (  i,j\right ) = \mathbb{1}\left [ \tilde{y}_{t}^{'}\left ( i \right ) \neq \tilde{y}_{t-1}^{'}\left ( j \right ) \right ]
\end{equation}
where $\mathbb{1}\left ( \cdot  \right )$ is the indicator function. $M_{p}\left (  i,j\right ) = 1$ when $i$th label in $\tilde{y}_{t}^{'}$ and $j$th label in $\tilde{y}_{t-1}^{'}$ are the same class, while $M_{p}\left (  i,j\right ) = 0$ otherwise; and $M_{n}$ vise versa. The value of cosine similarity map $S\in \mathbb{R}^{hw\times hw}$ corresponds to pixel $\left (  i,j\right )$ is determined by:
$
S\left ( i,j \right ) = \frac{f_{t}\left ( i \right )\cdot f_{t-1}\left ( j \right )}{\left\| f_{t}\left ( i \right )\right\|\times \left\| f_{t-1}\left ( j \right )\right\|}.
$
We then use the average cosine similarity of a pixel with all pixels in the previous frame belonging to the same or different class to represent its positive or negative affinity:

\begin{equation}
a_{p,f}\left ( i \right ) = \frac{\sum_{j=1}^{hw}S\left ( i,j \right )M_{p}\left ( i,j \right )}{\sum_{j=1}^{hw}M_{p}\left ( i,j \right )},\quad a_{n,f}\left ( i \right ) = \frac{\sum_{j=1}^{hw}S\left ( i,j \right )M_{n}\left ( i,j \right )}{\sum_{j=1}^{hw}M_{n}\left ( i,j \right )}
\end{equation}
where $a_{p,f},a_{n,f} \in \mathbb{R}^{h\times w}$ means the positive and negative map with the same size as the feature map. With simple up-sampling, we could obtain the final affinity maps $a_{p},a_{n} \in \mathbb{R}^{H\times W}$, indicating the positive and negative affinity of pixels in the current frame.

\subsection{Multi-Scale Supervision}
The feature map is first connected with a segmentation head generating the prediction $p$. Besides the standard cross entropy loss $\mathscr{L}^{CE} \left (p,\tilde{y}\right )= -\sum_{i}^{HW}\tilde{y}\left ( i \right )logp\left ( i \right )$, we applied a label corrected cross entropy loss $\mathscr{L}_{LC}^{CE} \left (p,\hat{y}\right )= -\sum_{i}^{HW}\hat{y}\left ( i \right )logp\left ( i \right )$ to train the network with pixel-level corrected labels. We further use two weight factors $\lambda _{I}$ and $\lambda _{V}$ to supervise the network in image and video levels. The specific descriptions are explained in the following sections.

\noindent \textbf{Pixel-Level Supervision.}
Inspired by the principle in Confident Learning~\cite{northcutt2021confident}, we use affinity maps to denote the confidence of labels. if a pixel $x\left ( i \right )$ in an image has both small enough positive affinity $a_{p}\left ( i \right )\leqslant t_{p}$ and large enough negative affinity $a_{n}\left ( i \right )\geqslant  t_{n}$, then its label $\tilde{y}\left ( i \right )$ can be suspected as noisy. The threshold $t_{p},t_{n}$ are obtained empirically by calculating the average positive and negative affinity, formulated as $t_{p} = \frac{1}{\left| A_{p}\right|}\sum_{a_{p}\in A_{p}}^{}\overline{a_{p}}$, $t_{n} = \frac{1}{\left| A_{n}\right|}\sum_{a_{n}\in A_{n}}^{}\overline{a_{n}}$, where $\overline{a_{p}},\overline{a_{n}}$ means the average value of positive and negative affinity over an image. The noisy pixels set can therefore be defined by:

\begin{equation}
\tilde{x} := \left\{ x\left ( i \right )\in x: a_{p}\left ( i \right )\leqslant t_{p}\right\}\bigcap \left\{ x\left ( i \right )\in x: a_{n}\left ( i \right )\geqslant  t_{n}\right\}.
\label{eq_noisy_map}
\end{equation}
Then we update the pixel-level label map $\hat{y}$ as:

\begin{equation}
\hat{y}(i) = \mathbb{1}\left (  x(i)\in \tilde{x}\right )p(i) + \mathbb{1}\left (  x(i)\notin \tilde{x}\right )\tilde{y}(i),
\end{equation}
where $p(i)$ is the prediction of network. Through this process, we only replace those pixels with both low positive affinity and large negative affinity.

\noindent \textbf{Image-Level Supervision.}
Even in the same video, different frames may contain different amounts of noisy labels. Hence, we first define the affinity confidence value as: $q = \overline{a_{p}} + 1 - \overline{a_{n}}$. The average affinity confidence value is therefore denoted as: $\bar{q} = t_{p} + 1 - t_{n}$. Finally, we define the image-level weight as:

\begin{equation}
\lambda_{I}  = e^{2(q - \bar{q})}.
\end{equation}
$\lambda_{I}>1$ if the sample has large affinity confidence and $\lambda_{I}<1$ otherwise, therefore enabling the network to concentrate more on the clean samples.

\noindent \textbf{Video-Level Supervision.}
We assign different weights to different videos such that the network can learn from more correct videos in the early stage. We first define the video affinity confidence as the average affinity confidence of all the frames: $q_{v} = \frac{1}{\left| V\right|}\sum_{x\in V}^{}q_{x}$. Supposing there are $N$ videos in total, we use $k\in \left\{ 1,2,\cdots ,N\right\}$ to represent the ranking of video affinity confidence from small to large, which means $k=1$ and $k=N$ denote the video with lowest and highest affinity confidence separately. Video-level weight is thus formulated as:

\begin{equation}
\lambda_{V} =\begin{cases}
\theta_{l}, & \text{ if } k<\frac{N}{3} \\
\theta_{l} + \frac{3k-N}{N}(\theta_{u} - \theta_{l}), & \text{ if } \frac{N}{3}\leqslant k\leqslant\frac{2N}{3}  \\
\theta_{u}, & \text{ if } k>\frac{2N}{3} 
\end{cases}
\end{equation}
where $\theta_{l}$ and $\theta_{u}$ are the preseted lower-bound and upper-bound of weight.

Combining the above-defined losses and weights, we obtain the final loss as: $\mathscr{L} = \lambda_{V} \lambda_{I} \mathscr{L}^{CE}+\mathscr{L}_{LC}^{CE}$, which supervise the network in a multi-scale manner. These losses and weights are enrolled in training after initialization in an order of video, image, and pixel enabling the network to enhance the robustness and generalization ability by concentrating on clean samples from rough to subtle.

\section{Experiments}

\subsection{Dataset Description and Experiment Settings}

\textbf{EndoVis 2018 Dataset and Noise Patterns.} EndoVis 2018 Dataset is from the MICCAI robotic instrument segmentation dataset\footnote{\url{https://endovissub2018-roboticscenesegmentation.grand-challenge.org/}} of endoscopic vision challenge 2018~\cite{allan20202018}. It is officially divided into 15 videos with 2235 frames for training and 4 videos with 997 frames for testing separately. The dataset contains 12 classes including different anatomy and robotic instruments. Each image is resized into $256 \times 320$ in pre-processing. To better simulate manual noisy annotations within a video, we first randomly select a ratio of $\alpha$ of videos and in each selected video, we divide all frames into several groups in a group of $3\sim 6$ consecutive frames. Then for each group of frames, we randomly apply dilation, erosion, affine transformation, or polygon noise to each class~\cite{li2021superpixel,zhang2020characterizing, xue2020cascaded, zhang2020robust}. We investigated our algorithms in several noisy settings with $\alpha$ being $\left\{ 0.3,0.5,0.8\right\}$. Some examples of data and noisy labels are shown in supplementary.

\noindent \textbf{Rat Colon Dataset.} For real-world noisy dataset, we have collected rat colon OCT images using 800nm ultra-high resolution endoscopic spectral domain OCT. We refer readers to~\cite{Yuan:22} for more details. Each centimeter of rat colon imaged corresponds to 500 images with 6 class layers of interest. We select 8 sections with 2525 images for training and 3 sections with 1352 images for testing. The labels of test set were annotated by professional endoscopists as ground truth while the training set was annotated by non-experts. Each image is resized into $256 \times 256$ in pre-processing. Some dataset examples are shown in supplementary.

\noindent \textbf{Implementation Details.}
We adopt Deeplabv3+~\cite{chen2018encoder} as our backbone network for fair comparison. The framework is implemented with PyTorch on two Nvidia 3090 GPUs. We adopt the Adam optimizer with an initial learning rate of $1e-4$ and weight decay of $1e-4$. Batch size is set to 4 with a maximum of 100 epochs for both Datasets. $\theta_{l}$ and $\theta_{u}$ are set to $0.4$ and $1$ separately. The video, image, and pixel level supervision are involved from the $16$th, $24$th, and $40$th epoch respectively. The segmentation performance is assessed by \emph{mIOU} and \emph{Dice} scores.

\subsection{Experiment Results on EndoVis 2018 Dataset}

\begin{table}[t]
\centering
\caption{Comparison of other methods and our models on EndoVis 2018 Dataset under different ratios of noise. The best results are \textbf{highlighted}.}
\scalebox{0.8}{
\begin{tabular}{m{1.4cm}<{\centering}|m{3cm}<{\centering}|m{1.5cm}<{\centering}|m{1.5cm}<{\centering}m{1.5cm}<{\centering}m{1.5cm}<{\centering}m{1.5cm}<{\centering}|m{1.5cm}<{\centering}}
\toprule
\multicolumn{1}{c|}{\multirow{2}{*}{Data}}   & \multicolumn{1}{c|}{\multirow{2}{*}{Method}}  & \multirow{2}{*}{\emph{mIOU}(\%) }              
    & \multicolumn{4}{c|}{Sequence \emph{mIOU}(\%)}   & \multicolumn{1}{c}{\multirow{2}{*}{\emph{Dice}(\%)}}  \\  \cline{4-7} & & & Seq 1 & Seq 2 & Seq 3 & Seq 4 & \\ \midrule
\multicolumn{1}{c|}{Clean} & Deeplabv3+~\cite{chen2018encoder}  & 53.98  & 54.07 & 51.46	& 72.35	& 38.02	& 64.30 \\ 
 \midrule
 \multicolumn{1}{c|}{\multirow{6}{*}{Noisy, $\alpha = 0.3$}} & Deeplabv3+~\cite{chen2018encoder} & 50.42	& 49.90	& 48.50	& 67.18	& 36.10	& 60.60 \\
 & STswin $\left (  22 '\right )$~\cite{jin2022exploring} & 50.29	& \textbf{49.96}	& 48.67	& 66.52	& 35.99	& 60.62  \\
 & RAUNet $\left (  19 '\right )$~\cite{ni2019raunet} & 50.36	& 44.97	& 48.06	& 68.90	& \textbf{39.53}	& 60.61  \\
 & JCAS $\left (  22 '\right )$~\cite{guo2022joint} & 48.65	& 48.77	& 46.60	& 64.83	& 34.39	& 58.97  \\
 & VolMin $\left (  21 '\right )$~\cite{pmlr-v139-li21l} & 47.64	& 45.60	& 45.31	& 64.01	& 35.63	& 57.42  \\
 & MS-TFAL(Ours) & \textbf{52.91}	& 49.48	& \textbf{51.60}	& \textbf{71.08}	& 39.52	& \textbf{62.91} \\ \midrule
\multicolumn{1}{c|}{\multirow{6}{*}{Noisy, $\alpha = 0.5$}} & Deeplabv3+~\cite{chen2018encoder} & 42.87	& 41.72	& 42.96	& 59.54	& 27.27	& 53.02 \\
 & STswin $\left (  22 '\right )$~\cite{jin2022exploring} & 44.48	& 40.78	& 45.22	& 60.50	& 31.45	& 54.99  \\
 & RAUNet $\left (  19 '\right )$~\cite{ni2019raunet} & 46.74	& 46.16	& 43.08	& 63.00	& \textbf{34.73}	& 57.44  \\
 & JCAS $\left (  22 '\right )$~\cite{guo2022joint} & 45.24	& 41.90	& 44.06	& 61.13	& 33.90	& 55.22  \\
 & VolMin $\left (  21 '\right )$~\cite{pmlr-v139-li21l} & 44.02	& 42.68 & 46.26	& 59.67	& 27.47	& 53.59  \\
 & MS-TFAL(Ours) & \textbf{50.34}	& \textbf{49.15}	& \textbf{50.17}	& \textbf{67.37}	& 34.67	& \textbf{60.50} \\ \midrule
\multicolumn{1}{c|}{\multirow{6}{*}{Noisy, $\alpha = 0.8$}}& Deeplabv3+~\cite{chen2018encoder}  & 33.35	& 27.57	& 35.69	& 45.30	& 24.86	& 42.22 \\
 & STswin $\left (  22 '\right )$~\cite{jin2022exploring} & 32.27	& 28.92	& 34.48	& 42.97	& 22.72	& 42.61  \\
 & RAUNet $\left (  19 '\right )$~\cite{ni2019raunet} & 33.25	& 30.23	& 34.95	& 44.99	& 22.88	& 43.67  \\
 & JCAS $\left (  22 '\right )$~\cite{guo2022joint} & 35.99	& 28.29	& 38.06	& 51.00	& 26.66	& 44.75  \\
 & VolMin $\left (  21 '\right )$~\cite{pmlr-v139-li21l} & 33.85	& 28.40	& 39.38	& 43.76	& 23.90	& 42.63  \\
 & MS-TFAL(Ours) & \textbf{41.36}	& \textbf{36.33}	& \textbf{41.65}	& \textbf{59.57}	& \textbf{27.88}	& \textbf{51.01} \\ \midrule \midrule
\multicolumn{1}{c|}{\multirow{5}{*}{Noisy, $\alpha = 0.5$}} & w/ V & 47.80	& 44.73	& 48.71	& 66.87	& 30.91	& 57.45 \\
& w/ V \& I  & 48.72	& 43.20	& 48.44	& 66.34	& \textbf{36.93}	& 58.54 \\
& Same frame  & 48.99	& 46.77	& 49.58	& 64.70	& 34.94	& 59.31 \\
& Any frame  & 48.69	& 46.25	& 48.92	& 65.56	& 34.08	& 58.89 \\
& MS-TFAL(Ours)  & \textbf{50.34}	& \textbf{49.15}	& \textbf{50.17}	& \textbf{67.37}	& 34.67	& \textbf{60.50}\\

\bottomrule
\end{tabular}}
\label{tab_main}
\end{table}

\begin{figure}[t!]
\centering
\includegraphics[width=1\linewidth]{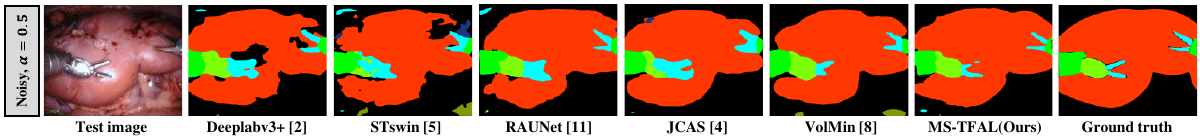}
\caption{Comparison of qualitative segmentation results on EndoVis18 Dataset.}
\label{fig:examples}
\end{figure}

Table~\ref{tab_main} presents the comparison results under different ratios of label noises. We evaluate the performance of backbone trained with clean labels, two state-of-the-art instrument segmentation network~\cite{jin2022exploring,ni2019raunet}, two noisy label learning techniques~\cite{guo2022joint,pmlr-v139-li21l}, backbone~\cite{chen2018encoder} and the proposed MS-TFAL. We re-implement~\cite{guo2022joint,pmlr-v139-li21l} with the same backbone~\cite{chen2018encoder} for a fair comparison. Compared with all other methods, MS-TFAL shows the minimum performance gap with the upper bound (Clean) for both \emph{mIOU} and \emph{Dice} scores under all ratios of noises demonstrating the robustness of our method. As noise increases, the performance of all baselines decreases significantly indicating the huge negative effect of noisy labels. It is noteworthy that when the noise ratio rises from 0.3 to 0.5 and from 0.5 to 0.8, our method only drops 2.57\% \emph{mIOU} with 2.41\% \emph{Dice} and 8.98\% \emph{mIOU} with 9.49\% \emph{Dice}, both are the minimal performance degradation, which further demonstrates the robustness of our method against label noise. In the extreme noise setting ($\alpha = 0.8$), our method achieves 41.36\% \emph{mIOU} and 51.01\% \emph{Dice} and outperforms second best method 5.37\% \emph{mIOU} and 6.26\% \emph{Dice}. As shown in Fig.~\ref{fig:examples}, we provide partial qualitative results indicating the superiority of MS-TFAL over other methods in the qualitative aspect. More qualitative results are shown in supplementary.

\noindent \textbf{Ablation Studies.} We further conduct two ablation studies on our multi-scale components and choice of frame for feature affinity under noisy dataset with $\alpha = 0.5$. With only video-level supervision (w/ V), \emph{mIOU} and \emph{Dice} are increased by 4.93\% and 4.43\% compared with backbone only. Then we apply both video and image level supervision (w/ V \& I) and gain an increase of 0.92\% \emph{mIOU} and 1.09\% \emph{Dice}. Pixel-level supervision is added at last forming the complete Multi-Scale Supervision results in another improvement of 1.62\% \emph{mIOU} and 1.96\% \emph{Dice} verifying the effectiveness in attenuating noisy label issues of individual components. For the ablation study of the choice of frame, we compared two different attempts with ours: conduct TFAL with the same frame and any frame in the dataset (Ours is adjacent frame). Results show that using adjacent frame has the best performance compared to the other two choices.

\begin{figure}[t!]
\centering
\includegraphics[width=0.9\linewidth]{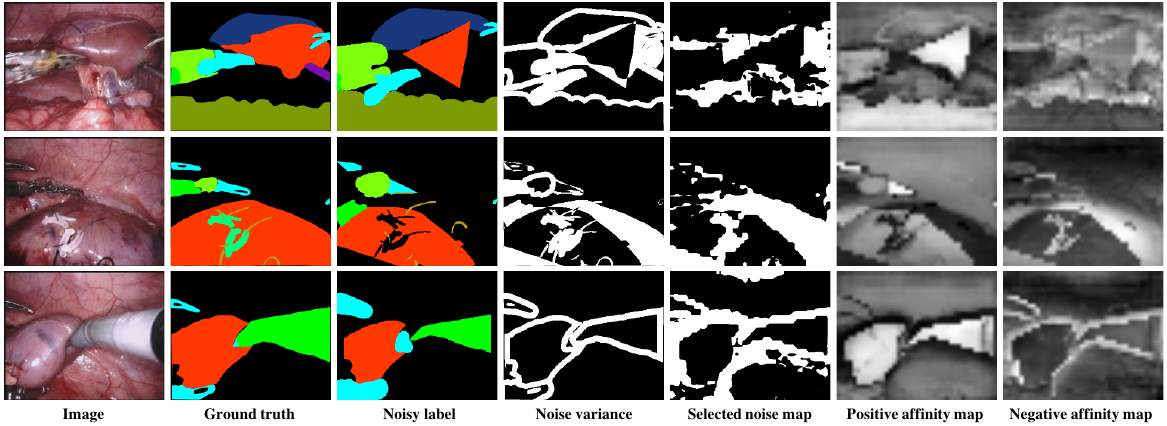}
\caption{Illustration of Noise variance and feature affinity. Selected noisy label (Fifth column) means the noise map selected with Equation~(\ref{eq_noisy_map}).}
\label{fig:Ex_TFAL}
\end{figure}

\noindent \textbf{Visualization of Temporal Affinity.} To prove the effectiveness of using affinity relation we defined to represent the confidence of label, we display comparisons between noise variance and selected noise map in Fig.~\ref{fig:Ex_TFAL}. Noise variance (Fourth column) represents the incorrect label map and the Selected noise map (Fifth column) denotes the noise map we select with Equation~(\ref{eq_noisy_map}). We can observe that the noisy labels we affirm have a high overlap degree with the true noise labels, which demonstrates the validity of our TFAL module.

\subsection{Experiment Results on Rat Colon Dataset}
The comparison results on real-world noisy Rat Colon Dataset are presented in Table~\ref{tab_OCT}. Our method outperforms other methods consistently on both \emph{mIOU} and \emph{Dice} scores, which verifies the superior robustness of our method on real-world label noise issues. Qualitative results are shown in supplementary.

\begin{table}[t]
\centering
\caption{Comparison of other methods and our models on Rat Colon Dataset.}
\scalebox{0.8}{
\begin{tabular}{m{1.7cm}<{\centering}|m{2.2cm}<{\centering}|m{1.9cm}<{\centering}|m{2cm}<{\centering}|m{1.9cm}<{\centering}|m{1.9cm}<{\centering}|m{2.6cm}<{\centering}}
\toprule
Method & Deeplabv3+~\cite{chen2018encoder} & STswin~\cite{jin2022exploring} & RAUNet~\cite{ni2019raunet} & JCAS~\cite{guo2022joint} & VolMin~\cite{pmlr-v139-li21l} & MS-TFAL(Ours) \\ \midrule
\emph{mIOU}(\%) & 68.46 & 68.21 & 68.24 & 68.15 & 68.81 & \textbf{71.05} \\ 
\emph{Dice}(\%) &75.25 & 77.70 & 77.39 & 77.50 & 77.89 & \textbf{80.17} \\ \bottomrule
\end{tabular}}
\label{tab_OCT}
\end{table}

\section{Discussion and Conclusion}

In this paper, we propose a robust MS-TFAL framework to resolve noisy label issues in medical video segmentation. Different from previous methods, we first introduce the novel TFAL module to use affinity between pixels from adjacent frames to represent the confidence of label. We further design MSS framework to supervise the network from multiple perspectives. Our method can not only identify noise in labels, but also correct them in pixel-wise with rich temporal consistency. Extensive experiments under both synthetic and real-world label noise data demonstrate the excellent noise resilience of MS-TFAL.

\subsection*{Acknowledgements}
This work was supported by Hong Kong Research Grants Council (RGC) Collaborative Research Fund (C4026-21G), General Research Fund (GRF 14211420 \& 14203323),  Shenzhen-Hong Kong-Macau Technology Research Programme (Type C) STIC Grant SGDX20210823103535014 (202108233000303).

\bibliography{mybib}{}

\begin{thebibliography}{10}
\providecommand{\url}[1]{\texttt{#1}}
\providecommand{\urlprefix}{URL }
\providecommand{\doi}[1]{https://doi.org/#1}

\bibitem{allan20202018}
Allan, M., Kondo, S., Bodenstedt, S., Leger, S., Kadkhodamohammadi, R., Luengo,
  I., Fuentes, F., Flouty, E., Mohammed, A., Pedersen, M., et~al.: 2018 robotic
  scene segmentation challenge. arXiv preprint arXiv:2001.11190  (2020)

\bibitem{chen2018encoder}
Chen, L.C., Zhu, Y., Papandreou, G., Schroff, F., Adam, H.: Encoder-decoder
  with atrous separable convolution for semantic image segmentation. In:
  Proceedings of the European conference on computer vision (ECCV). pp.
  801--818 (2018)

\bibitem{guo2021metacorrection}
Guo, X., Yang, C., Li, B., Yuan, Y.: Metacorrection: Domain-aware meta loss
  correction for unsupervised domain adaptation in semantic segmentation. In:
  Proceedings of the IEEE/CVF Conference on Computer Vision and Pattern
  Recognition. pp. 3927--3936 (2021)

\bibitem{guo2022joint}
Guo, X., Yuan, Y.: Joint class-affinity loss correction for robust medical
  image segmentation with noisy labels. In: Medical Image Computing and
  Computer Assisted Intervention--MICCAI 2022: 25th International Conference,
  Singapore, September 18--22, 2022, Proceedings, Part IV. pp. 588--598.
  Springer (2022)

\bibitem{jin2022exploring}
Jin, Y., Yu, Y., Chen, C., Zhao, Z., Heng, P.A., Stoyanov, D.: Exploring
  intra-and inter-video relation for surgical semantic scene segmentation. IEEE
  Transactions on Medical Imaging  \textbf{41}(11),  2991--3002 (2022)

\bibitem{karimi2020deep}
Karimi, D., Dou, H., Warfield, S.K., Gholipour, A.: Deep learning with noisy
  labels: Exploring techniques and remedies in medical image analysis. Medical
  image analysis  \textbf{65},  101759 (2020)

\bibitem{li2021superpixel}
Li, S., Gao, Z., He, X.: Superpixel-guided iterative learning from noisy labels
  for medical image segmentation. In: Medical Image Computing and Computer
  Assisted Intervention--MICCAI 2021: 24th International Conference,
  Strasbourg, France, September 27--October 1, 2021, Proceedings, Part I 24.
  pp. 525--535. Springer (2021)

\bibitem{pmlr-v139-li21l}
Li, X., Liu, T., Han, B., Niu, G., Sugiyama, M.: Provably end-to-end
  label-noise learning without anchor points. In: Meila, M., Zhang, T. (eds.)
  Proceedings of the 38th International Conference on Machine Learning.
  Proceedings of Machine Learning Research, vol.~139, pp. 6403--6413. PMLR
  (18--24 Jul 2021), \url{https://proceedings.mlr.press/v139/li21l.html}

\bibitem{LITJENS201760}
Litjens, G., Kooi, T., Bejnordi, B.E., Setio, A.A.A., Ciompi, F., Ghafoorian,
  M., {van der Laak}, J.A., {van Ginneken}, B., Sánchez, C.I.: A survey on
  deep learning in medical image analysis. Medical Image Analysis  \textbf{42},
   60--88 (2017). \doi{https://doi.org/10.1016/j.media.2017.07.005},
  \url{https://www.sciencedirect.com/science/article/pii/S1361841517301135}

\bibitem{liu2021s}
Liu, L., Zhang, Z., Li, S., Ma, K., Zheng, Y.: S-cuda: self-cleansing
  unsupervised domain adaptation for medical image segmentation. Medical Image
  Analysis  \textbf{74},  102214 (2021)

\bibitem{ni2019raunet}
Ni, Z.L., Bian, G.B., Zhou, X.H., Hou, Z.G., Xie, X.L., Wang, C., Zhou, Y.J.,
  Li, R.Q., Li, Z.: Raunet: Residual attention u-net for semantic segmentation
  of cataract surgical instruments. In: International Conference on Neural
  Information Processing. pp. 139--149. Springer (2019)

\bibitem{northcutt2021confident}
Northcutt, C., Jiang, L., Chuang, I.: Confident learning: Estimating
  uncertainty in dataset labels. Journal of Artificial Intelligence Research
  \textbf{70},  1373--1411 (2021)

\bibitem{dlsurvey}
Shen, D., Wu, G., Suk, H.I.: Deep learning in medical image analysis. Annual
  Review of Biomedical Engineering  \textbf{19}(1),  221--248 (2017).
  \doi{10.1146/annurev-bioeng-071516-044442},
  \url{https://doi.org/10.1146/annurev-bioeng-071516-044442}, pMID: 28301734

\bibitem{shi2021distilling}
Shi, J., Wu, J.: Distilling effective supervision for robust medical image
  segmentation with noisy labels. In: Medical Image Computing and Computer
  Assisted Intervention--MICCAI 2021: 24th International Conference,
  Strasbourg, France, September 27--October 1, 2021, Proceedings, Part I 24.
  pp. 668--677. Springer (2021)

\bibitem{xu2022anti}
Xu, Z., Lu, D., Luo, J., Wang, Y., Yan, J., Ma, K., Zheng, Y., Tong, R.K.Y.:
  Anti-interference from noisy labels: Mean-teacher-assisted confident learning
  for medical image segmentation. IEEE Transactions on Medical Imaging
  \textbf{41}(11),  3062--3073 (2022)

\bibitem{xue2020cascaded}
Xue, C., Deng, Q., Li, X., Dou, Q., Heng, P.A.: Cascaded robust learning at
  imperfect labels for chest x-ray segmentation. In: Medical Image Computing
  and Computer Assisted Intervention--MICCAI 2020: 23rd International
  Conference, Lima, Peru, October 4--8, 2020, Proceedings, Part VI 23. pp.
  579--588. Springer (2020)

\bibitem{Yuan:22}
Yuan, W., Feng, Y., Chen, D., Gharibani, P., Chen, J.D.Z., Yu, H., Li, X.: In
  vivo assessment of inflammatory bowel disease in rats with
  ultrahigh-resolution colonoscopic oct. Biomed. Opt. Express  \textbf{13}(4),
  2091--2102 (Apr 2022). \doi{10.1364/BOE.453396},
  \url{https://opg.optica.org/boe/abstract.cfm?URI=boe-13-4-2091}

\bibitem{zhang2020characterizing}
Zhang, M., Gao, J., Lyu, Z., Zhao, W., Wang, Q., Ding, W., Wang, S., Li, Z.,
  Cui, S.: Characterizing label errors: confident learning for noisy-labeled
  image segmentation. In: Medical Image Computing and Computer Assisted
  Intervention--MICCAI 2020: 23rd International Conference, Lima, Peru, October
  4--8, 2020, Proceedings, Part I 23. pp. 721--730. Springer (2020)

\bibitem{zhang2020robust}
Zhang, T., Yu, L., Hu, N., Lv, S., Gu, S.: Robust medical image segmentation
  from non-expert annotations with tri-network. In: Medical Image Computing and
  Computer Assisted Intervention--MICCAI 2020: 23rd International Conference,
  Lima, Peru, October 4--8, 2020, Proceedings, Part IV 23. pp. 249--258.
  Springer (2020)

\end{thebibliography}
\bibliographystyle{splncs04}

\newpage
\section{Supplementary}

\begin{figure}[thbp!]
\centering
\includegraphics[width=0.81\linewidth]{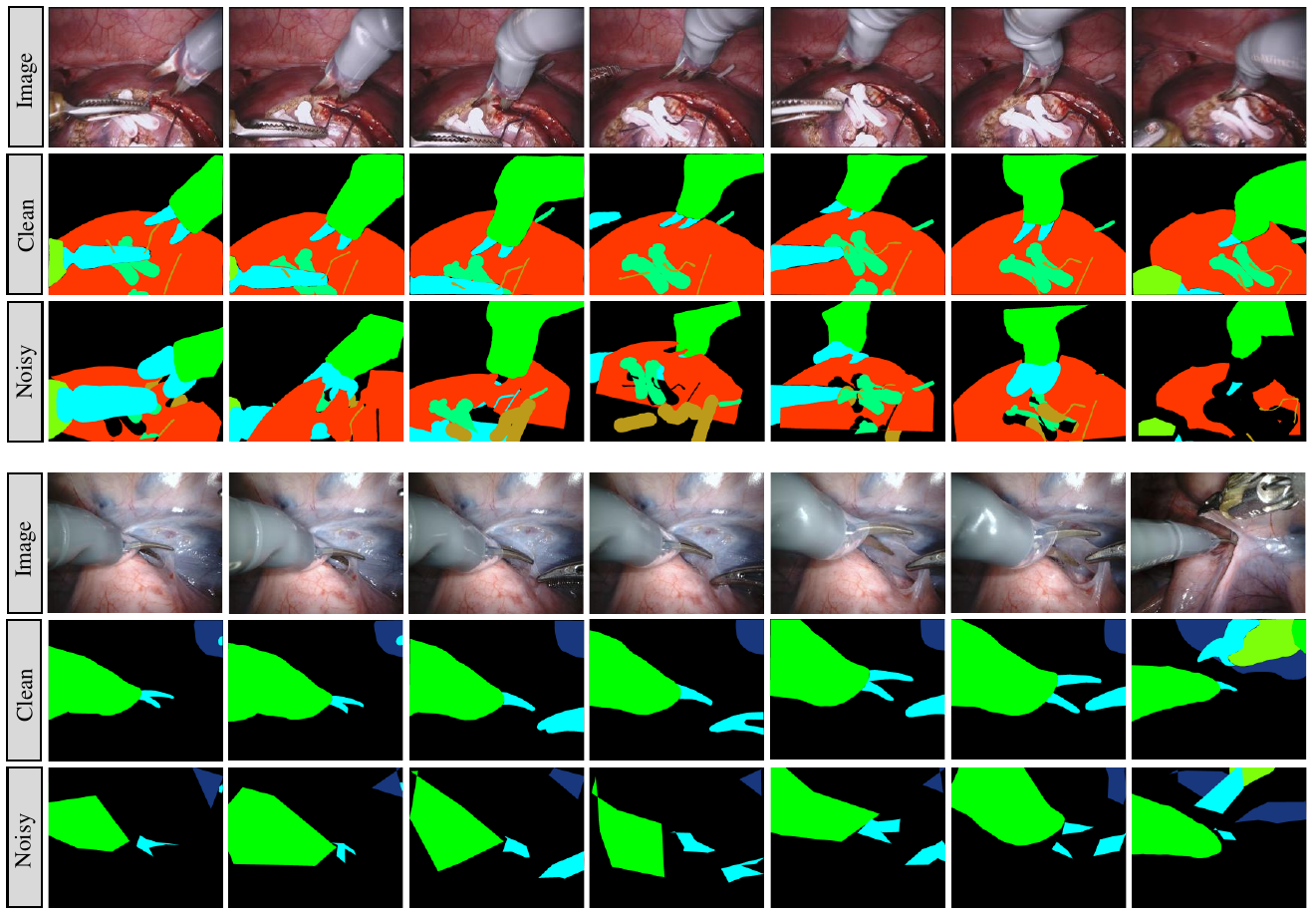}
\caption{Examples of EndoVis 2018 Dataset\cite{allan20202018} including images , clean and noisy labels.}
\label{fig:Sup 1L}
\end{figure}

\begin{figure}[thbp!]
\centering
\includegraphics[width=0.81\linewidth]{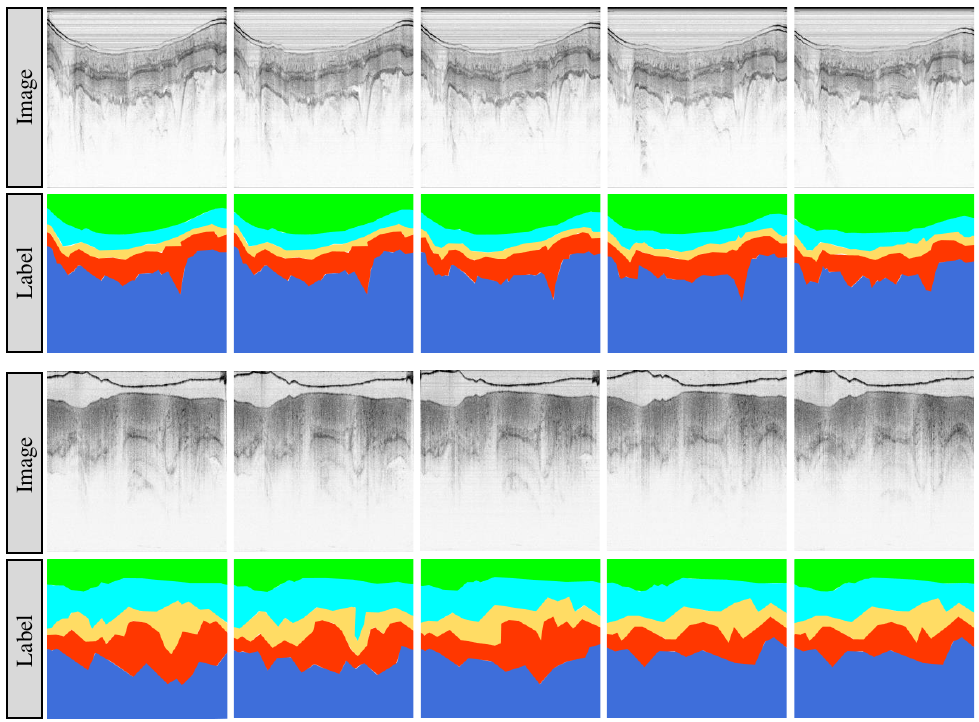}
\caption{Examples of Rat Colon Dataset.}
\label{fig:Sup 2L}
\end{figure}

\begin{figure}[thbp!]
\centering
\includegraphics[width=0.81\linewidth]{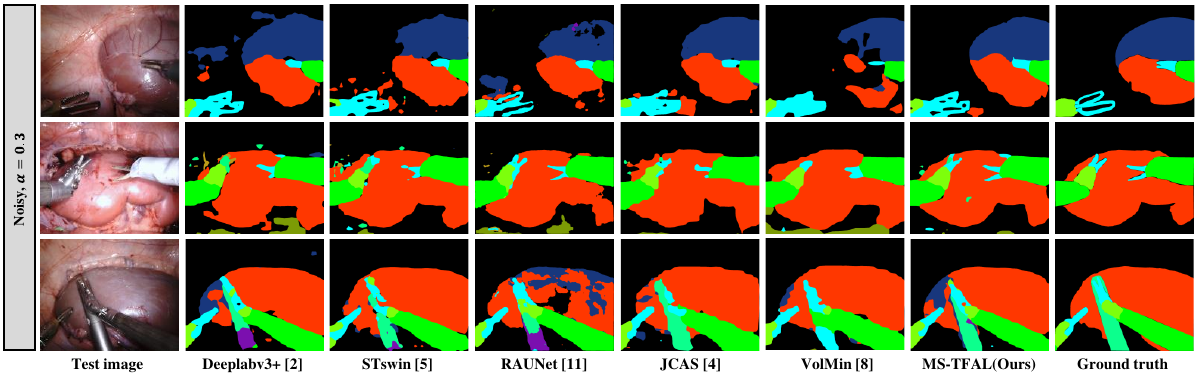}
\caption{Qualitative segmentation results on EndoVis 2018 Dataset ($\alpha = 0.3$).}
\label{fig:Sup 3L}
\end{figure}

\begin{figure}[thbp!]
\centering
\includegraphics[width=0.81\linewidth]{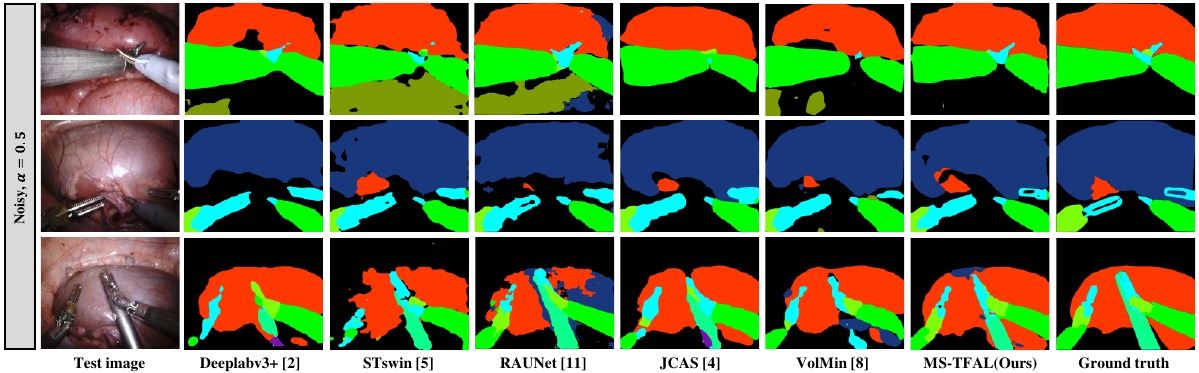}
\caption{Qualitative segmentation results on EndoVis 2018 Dataset ($\alpha = 0.5$).}
\label{fig:Sup 4L}
\end{figure}

\begin{figure}[thbp!]
\centering
\includegraphics[width=0.81\linewidth]{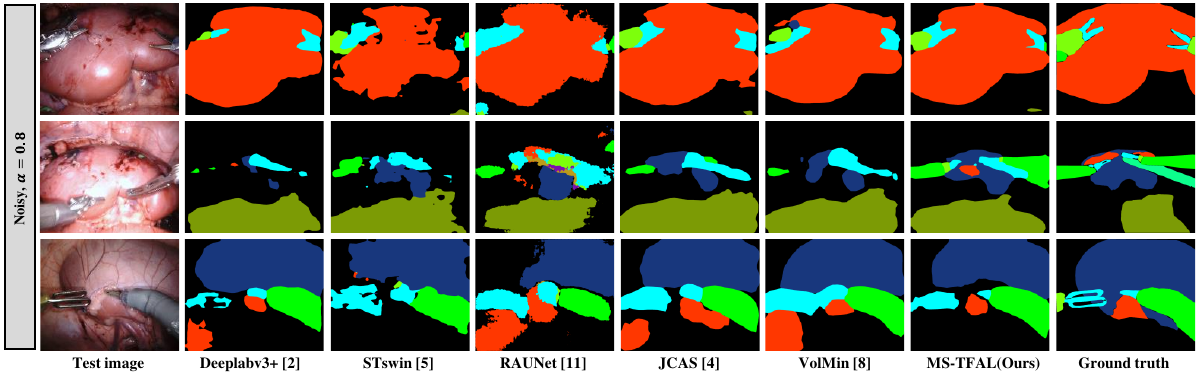}
\caption{Qualitative segmentation results on EndoVis 2018 Dataset ($\alpha = 0.8$).}
\label{fig:Sup 5L}
\end{figure}

\begin{figure}[thbp!]
\centering
\includegraphics[width=0.81\linewidth]{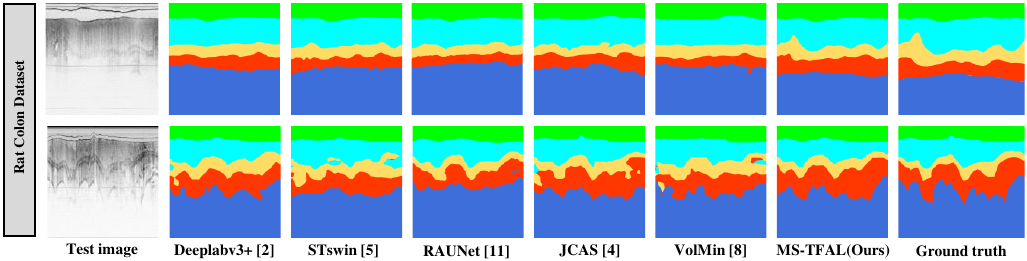}
\caption{Qualitative segmentation results on Rat Colon Dataset.}
\label{fig:Sup 6L}
\end{figure}
\end{document}